\pdfoutput=1

\documentclass[11pt]{article}

\usepackage[]{acl}
\usepackage[export]{adjustbox}
\usepackage{times}
\usepackage{latexsym}
\usepackage{afterpage}
\usepackage[T1]{fontenc}

\usepackage[utf8]{inputenc}
\usepackage[greek,english]{babel}
\usepackage{comment}
\usepackage{microtype}

\usepackage{enumitem}

\usepackage{inconsolata}
\usepackage{longtable}
\usepackage{booktabs}
\usepackage{multirow}
\usepackage{graphicx}
\usepackage{textgreek}

\usepackage[normalem]{ulem}

\usepackage{tikz}
\usetikzlibrary{calc,matrix}
\newcommand{\hnode}[1]{|(#1)| #1}

\makeatletter
\tikzset{
  arm angleA/.initial={0},
  arm angleB/.initial={0},
  arm lengthA/.initial={0mm},
  arm lengthB/.initial={0mm},
  arm length/.style={%
    arm lengthA=#1,
    arm lengthB=#1,
  },
  arm/.style={
    to path={%
      (\tikztostart) -- ++(\pgfkeysvalueof{/tikz/arm angleA}:\pgfkeysvalueof{/tikz/arm lengthA}) -- ($(\tikztotarget)+(\pgfkeysvalueof{/tikz/arm angleB}:\pgfkeysvalueof{/tikz/arm lengthB})$) -- (\tikztotarget)
    }
  },
  expand/.code={%
    \let\pgf@matrix@compute@origin=\pgf@matrix@compute@origin@expand
    \let\pgf@matrix@cont=\pgf@matrix@cont@expand%
    \let\pgf@matrix@cell@cont=\pgf@matrix@cell@cont@expand
  },
  expand width/.initial={100pt}, 
}

\def\ex@minwidth{100pt}%
\let\pgf@matrix@compute@origin@orig=\pgf@matrix@compute@origin
\def\pgf@matrix@compute@origin@expand{%
  \pgf@matrix@compute@origin@orig
  \pgfmathsetmacro{\ex@width}{%
    \csname pgf@matrix@minx\the\pgf@matrix@numberofcolumns\endcsname -
    \csname pgf@matrix@minx1\endcsname +
    \csname pgf@matrix@maxx\the\pgf@matrix@numberofcolumns\endcsname +
    \csname pgf@matrix@maxx1\endcsname +
2*\pgfkeysvalueof{/pgf/inner xsep}
  }
  \pgfmathsetmacro{\ex@extra}{max(0,(\pgfkeysvalueof{/tikz/expand width} - \ex@width)/(\pgf@matrix@numberofcolumns - 1))}%
  {%
    \c@pgf@counta=1\relax%
    \advance\pgf@matrix@numberofcolumns by 1\relax
    \loop%
    \ifnum\c@pgf@counta<\pgf@matrix@numberofcolumns\relax%
    \pgfmathparse{\csname pgf@matrix@minx\the\c@pgf@counta\endcsname + (\c@pgf@counta - 1) * \ex@extra}%
      \expandafter\xdef\csname pgf@matrix@minx\the\c@pgf@counta\endcsname{\pgfmathresult pt}%
      \advance\c@pgf@counta by1\relax%
    \repeat%
  }%
}
\def\pgf@matrix@cont@expand{%
    \setbox\pgf@matrix@box=\hbox\bgroup\vbox\bgroup%
  \pgfmathparse{\pgfkeysvalueof{/tikz/expand width} - 2*\pgfkeysvalueof{/pgf/inner xsep}}%
    \halign to \pgfmathresult pt\bgroup%
    \pgf@matrix@init@row%
    \pgf@matrix@step@column%
    {%
      \pgf@matrix@startcell%
      ##%
      \pgf@matrix@endcell%
    }%
    \tabskip=0pt\relax
    &%
    ##\pgf@matrix@padding&&%
    ##%
    \tabskip=0pt plus 1fil\relax
    &%
    \pgf@matrix@step@column%
    {%
      \pgf@matrix@startcell%
      ##%
      \pgf@matrix@endcell%
    }%
    \tabskip=0pt\relax
    &%
    ##\pgf@matrix@padding%
    \cr%
}

\def\pgf@matrix@cell@cont@expand[#1]{%
  \ifnum\pgfmatrixcurrentcolumn<\pgf@matrix@numberofcolumns%
  \else%
  {%
    \global\pgf@matrix@fixedfalse%
    \pgf@x=0pt%
    \pgf@matrix@addtolength{\pgf@x}{\pgfmatrixcolumnsep}%
    \pgf@matrix@addtolength{\pgf@x}{#1}%
    \ifpgf@matrix@fixed%
      \expandafter\pgfutil@g@addto@macro\csname pgf@matrix@column@finish@\the\pg
fmatrixcurrentcolumn\endcsname%
        {\global\pgf@picmaxx=0pt}%
    \fi%
    \advance\pgfmatrixcurrentcolumn by1\relax 
    \expandafter\xdef\csname pgf@matrix@column@sep@\the\pgfmatrixcurrentcolumn\endcsname{\the\pgf@x}%
    \ifpgf@matrix@fixed%
      \expandafter\gdef\csname pgf@matrix@column@finish@\the\pgfmatrixcurrentcolumn\endcsname{\global\pgf@picminx=0pt}%
    \else%
      \expandafter\global\expandafter\let\csname pgf@matrix@column@finish@\the\pgfmatrixcurrentcolumn\endcsname=\pgfutil@empty%
    \fi%
  }%
  \fi%
  &\pgf@matrix@correct@calltrue&\pgf@matrix@correct@calltrue&%
}%

\makeatother

\usepackage{makecell}

\usepackage{mathtools}
\usepackage{blkarray, bigstrut} %

\usepackage{arydshln}
\makeatletter
\def\adl@drawiv#1#2#3{%
        \hskip.5\tabcolsep
        \xleaders#3{#2.5\@tempdimb #1{1}#2.5\@tempdimb}%
                #2\z@ plus1fil minus1fil\relax
        \hskip.5\tabcolsep}
\newcommand{\cdashlinelr}[1]{%
  \noalign{\vskip\aboverulesep
           \global\let\@dashdrawstore\adl@draw
           \global\let\adl@draw\adl@drawiv}
  \cdashline{#1}
  \noalign{\global\let\adl@draw\@dashdrawstore
           \vskip\belowrulesep}}
\makeatother

\usepackage{tikz}
\usetikzlibrary{calc}
\usetikzlibrary{decorations.pathmorphing}
\usetikzlibrary{shapes,arrows,chains}
\usetikzlibrary{shapes.geometric}
\usetikzlibrary{patterns}
\usetikzlibrary{positioning}
\tikzset{>=latex}
\usepackage{pgfplots}
\usepackage{relsize}

\usepackage{array}

\newcolumntype{P}[1]{>{\centering\arraybackslash}p{#1}}

\usepackage{makecell}

\title{A Morphologically-Aware Dictionary-based Data Augmentation Technique for Machine Translation of Under-Represented Languages}

\author{Md Mahfuz Ibn Alam$^\alpha$ \qquad Sina Ahmadi$^{\alpha,\beta}$ \qquad Antonios Anastasopoulos$^{\alpha,\gamma}$ \\
        $^\alpha$Department of Computer Science, George Mason University \hfill $^{\beta}$University of Zurich \\ 
        $^\gamma$Archimedes AI Research Unit, RC Athena, Greece\\
        \texttt{\{malam21,sahmad46,antonis\}@gmu.edu}}

\begin{document}
\maketitle
\begin{abstract}
The availability of parallel texts is crucial to the performance of machine translation models. However, most of the world's languages face the predominant challenge of data scarcity. In this paper, we propose strategies to synthesize parallel data relying on morpho-syntactic information and using bilingual lexicons along with a small amount of \emph{seed} parallel data. Our methodology adheres to a \textit{realistic} scenario backed by the small parallel seed data. It is linguistically informed, as it aims to create augmented data that is more likely to be grammatically correct. 
    We analyze how our synthetic data can be combined with raw parallel data and demonstrate a consistent improvement in performance in our experiments on 14 languages (28 English$\leftrightarrow$X pairs) ranging from well- to very low-resource ones. Our method leads to improvements even when using only five seed sentences and a bilingual lexicon.\footnote{Data and code will be publicly released upon acceptance.}
\end{abstract}

\section{Introduction}

\begin{figure}[t]
\centering

\begin{tikzpicture}

    \matrix[column sep=0em,matrix of nodes] (Kurmanji) {
    \hnode{Ew} & \hnode{gîtarê} & \hnode{pir}  & \hnode{baş} & \hnode{lê} & \hnode{dide}\\
    };
    \path (Kurmanji.east);
    \pgfgetlastxy{\Frrx}{\Frry}%
    \path (Kurmanji.west);
    \pgfgetlastxy{\Frlx}{\Frly}%
    \pgfmathsetmacro{\Frwidth}{\Frrx - \Frlx}%
    \path (Kurmanji) ++(-.25in,.8in) node[matrix,column sep=0em,matrix of nodes,expand,expand width={\Frwidth pt}] (English) {
    \hnode{He} & \hnode{plays} & \hnode{the} & \hnode{guitar} & \hnode{very} & \hnode{well} \\
    };
    \begin{scope}[every path/.style={line width=4pt,white,double=black},every to/.style={arm}, arm angleA=-90, arm angleB=90, arm length=5mm]
    \draw (He) to (Ew);
    \draw (very) to (pir);
    \draw (the) to (gîtarê);
    \draw (plays) -- ++(0,-.4in) coordinate (p1) {}
     (p1) to[arm lengthA=0pt] (lê)
     (p1) to[arm lengthA=0pt] (dide);
    \draw (well) to (baş);
    \draw [magenta] (guitar) to (gîtarê);
    \end{scope}

\end{tikzpicture}

\underline{Analyze}: \hfill guitar = [\textsc{n;acc;sg}, lemma=\textsc{guitar}]

\underline{Replace}: \hfill \textsc{guitar} $\leftarrow$ \textsc{flower}, \textsc{flower}= \textsc{gul}

\underline{Generate}: \hfill[\textsc{n;acc;sg}, lemma=\textsc{gul}]$\rightarrow$ \textit{gulê}

\begin{tikzpicture}

    \matrix[column sep=0em,matrix of nodes] (Kurmanji) {
    \hnode{He} & \hnode{plays} & \hnode{the} & \hnode{flower}  & \hnode{very} & \hnode{well}\\
    };
    \path (Kurmanji.east);
    \pgfgetlastxy{\Frrx}{\Frry}%
    \path (Kurmanji.west);
    \pgfgetlastxy{\Frlx}{\Frly}%
    \pgfmathsetmacro{\Frwidth}{\Frrx - \Frlx}%
    \path (Kurmanji) ++(-.25in,.8in) node[matrix,column sep=0em,matrix of nodes,expand,expand width={\Frwidth pt}] (English) {
    \hnode{Ew} & \hnode{gulê} & \hnode{pir} & \hnode{baş} & \hnode{lê} & \hnode{dide} \\
    };
    \begin{scope}[every path/.style={line width=4pt,white,double=black},every to/.style={arm}, arm angleA=-90, arm angleB=90, arm length=5mm]
    \draw (Ew) to (He) ;
    \draw (pir) to (very);
    \draw (gulê) to (the);
    \draw (lê) to (plays);
    \draw (dide) to (plays);
    \draw (baş) to (well);
    \draw [cyan] (gulê) to (flower);
    \end{scope}

\end{tikzpicture}
\vspace*{-4mm}
\caption{A schema of our approach. After aligning `guitar' (in English) and `\textit{gîtarê}' (in Kurmanji Kurdish), the new word `flower' is randomly selected to replace `guitar' and its translation `\textit{gul}' in a bilingual dictionary is inflected according to its morphological features as `\textit{gulê}'. Small caps refer to lemmata.}
\label{fig:schema_technique}

\end{figure}
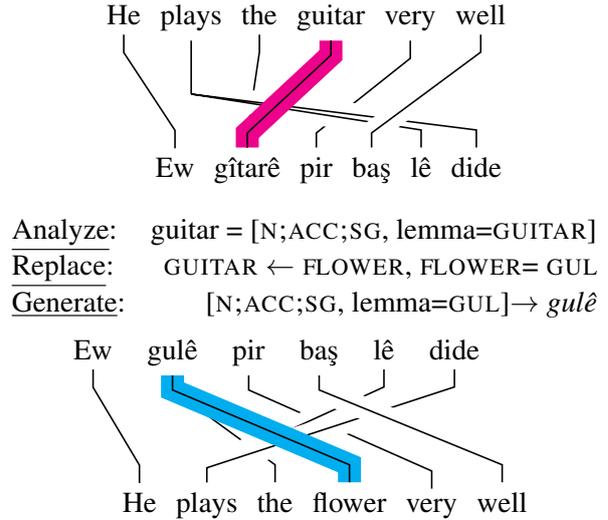

One of the major challenges in machine translation (MT) is the lack of parallel data for most of the world's languages. Traditional approaches \cite{wu-etal-2008-domain,DBLP:journals/corr/MikolovLS13} used to rely on dictionaries and linguistic knowledge for MT. One of the naive ways to use dictionaries for MT is to translate by looking up words of a source sentence in a bilingual lexicon and replacing their corresponding translations in the target language. 
However, this approach has certain shortcomings \cite{wang-etal-2022-expanding}. Firstly, the coverage of translations depends on the size and comprehensiveness of the lexicon, which can result in incomplete translations and code-mixed versions of the source and target languages. The translated sentences may also not adhere to the target language's grammatical rules or word order. Furthermore, most dictionaries operate at the lemma level, posing challenges for morphologically-rich languages. Therefore, \textit{solely} relying on dictionaries is not a viable solution for low-resource languages.

In recent approaches to MT that mainly rely on encoder-decoder networks like transformers \cite{DBLP:journals/corr/VaswaniSPUJGKP17}, the ideal scenario is to train an MT model on a large parallel corpus. Creating a parallel corpus for a given language, however, requires linguistic and technical expertise lacking for under-resourced languages and is also a costly and time-consuming task. To remedy this, recent studies in natural language processing (NLP) focus on unsupervised methods based on monolingual data ~\cite{DBLP:conf/acl/SennrichHB16,DBLP:conf/iclr/LampleCDR18}, back-translation ~\cite{DBLP:conf/emnlp/EdunovOAG18,edunov2018understanding}, other data augmentation techniques ~\cite{DBLP:conf/emnlp/Sanchez-Cartagena21}, or fine-tune pre-trained models to adapt to a different language, domain, or dialect ~\cite{DBLP:conf/emnlp/BapnaF19}. Therefore, the usage of dictionaries is largely under-studied, even though they are still practically in use~\cite{DBLP:journals/corr/abs-2004-02577,DBLP:conf/acl/SennrichHB16a}.

In this paper, we put forward a dictionary-based approach akin to early dictionary-based MT systems \cite{tyers-2009-rule,koehn-knight-2002-learning,koehn-knight-2001-knowledge,sanchez-cartagena-etal-2011-integrating} yet more sophisticated as it relies on the morpho-syntactic analysis of words to generate a parallel corpus synthetically. As illustrated in Figure~\ref{fig:schema_technique}, our approach consists of four components: alignment, analysis, replacement, and generation. Given a small set of parallel text as seed data, we first retrieve possible word-level translation pairs in the source and target languages as in `guitar' and `\textit{gîtar}' in English and Kurmanji Kurdish, respectively. We then morphologically analyze the source words in the translation pairs, e.g., `guitar' is a singular noun in the accusative case in the example. With the morphological features of a word in the source sentence, we can now sample a word from a bilingual dictionary with the same morphological features, e.g., \textit{gulê}, and "plug" it into our sentence to generate a new sentence pair synthetically. As such, the synthetically-generated sentences are likely to create new grammatically-sound translations.

To summarize, the contributions of our work are three-fold:

\begin{itemize}[noitemsep,nolistsep]
    \item We propose a morphologically-informed replacement method to create a new synthetic sentence.
    \item We show that this synthetic parallel data helps improve the MT system's quality when mixed with real parallel data.
    \item We also demonstrate the effectiveness of our method in extremely data-scarce scenarios, where as little as five parallel seed sentences are rendered useful with our approach.
\end{itemize}

Note that we will interchange the terms ``dictionary'' and ``bilingual lexicon'' throughout the paper for readability reasons.

\section{Method}

Our method requires a small parallel dataset called seed data containing sentences in the source and target languages to create synthetic parallel data. Our approach consists of three components. We first prepare data by tokenizing sentences and obtaining word-level alignment between the parallel sentences. This step is completed by morphologically analyzing aligned word pairs. Then, we replace words considering the morphological features in the augmentation component and filter the synthetic sentences using language models. Finally, we build MT systems in different settings varying the number of synthetic sentences.


\subsection{Analysis}

\paragraph{Alignment}
\label{par:Ali}
We perform word alignment to our seed data, identifying the relationship between words in the seed sentence. This is necessary for knowing which words are translations of each other. If we replace a word in the source sentence, the aligned target word of the target sentence must also be replaced to reflect the changes.

\paragraph{Morphological Tagging}
\label{par:Tag}
We analyze the entries in the bilingual lexicon's source side words to facilitate the data augmentation process. This way, we can categorize entries based on morphological features and find the part-of-speech (POS) tags, e.g., \textsc{adj}, of our bilingual lexicon's source side words.


\paragraph{Word-pair Selection}
We randomly choose words from the source side, here in English, for each seed sentence. We refer to Figure~\ref{fig:schema_technique} as our example where we generate the morphological feature and POS tag for the given word ``guitar''. We also find the translation of ``guitar'' in the seed sentence's target side. Here, that word is ``gîtarê", which we get from the alignment in §\ref{par:Ali}. We find the morphological feature and POS tag of ``gîtarê'' too.

\subsection{Augmentation}
We introduce two different approaches for the augmentation of the seed sentences:
\paragraph{Morphologically-Informed}

\begin{enumerate}
    \item Referring to Figure~\ref{fig:schema_technique},  we first replace ``guitar'' with another random word, e.g. ``flower'', having identical morphological features created in §\ref{par:Tag}. As such, a new sentence is synthetically created as ``\textit{He plays the flower very well}''. It should be noted that this procedure does not consider the semantic relevance of the candidate word. In other words, it may yield nonsensical sentences yet morpho-syntactically valid.
    
    \item Then, we replace ``\textit{gîtarê}'' with the translation of ``flower'' being ``\textit{gul}'' in Kurmanji Kurdish from the bilingual lexicon. It is worth mentioning that we use PanLex dictionaries\footnote{\url{https://panlex.org/snapshot}} where some of the entries are not in the lemma form. Therefore, we also lemmatize the retrieved word form in the dictionary, i.e., ``\textit{gul}'' to mitigate the impact of the inaccuracy of the lexicographic data. 
    
    \item Last, we perform morphological inflection where a lemma is inflected based on morphological features of the word that will be substituted, i.e., `\textit{gîtar}'. Doing this, we create a new sentence where the randomly selected word in the dictionary `\textit{gul}' appears grammatically and morphologically correct as `\textit{gulê}'.  We do this to guarantee that the new word follows the correct morphological features. Thus creating a synthetic target translation ``\textit{Ew gulê pir baş lê dide}'' of the synthetic source sentence ``He plays the flower very well''.
\end{enumerate}

\paragraph{Naive}

In contrast to the previous augmentation technique where we consider the morphological features, we carry out a naive random word replacement approach where only the POS tag is identical, without lemmatizing or inflecting the word based on the sentence. For instance in Figure~\ref{fig:schema_technique}, a synthetic sentence created this way would be ``He plays the flower very well'' and its generated translation ``\textit{Ew gul pir baş lê dide}''. Here ``\textit{gul}'' has not been converted into `\textit{gulê}'. In the \emph{Morphologically-Informed} setup, we preserve the morphological information of the word we change, thus making the synthetic data more likely to be grammatically correct. In this \emph{naive} approach, on the other hand, we lose this information.


\subsection{Filtering with LMs}
We create synthetic sentences for each seed sentence with the above augmentation approaches. Given that the synthetic sentences may not be meaningful, e.g., ``He plays the flower very well'', we also incorporate information from a language model (LM) by estimating the perplexity ($ppl$) of the synthetic sentences:
\[
    ppl(x) = exp\{-1/t \sum_{i}^t log p_{\theta}(x_i|x_{<i}) \},
\]
Where the probability of a sentence of length $t$ containing words $x$ existing in the LM. The lower the perplexity is, the more natural the sentence is. We filter the augmented sentences using the LM and rank them based on the perplexity scores to pick the sentences with the correct context. This step yields sentences more likely to appear with the lowest perplexity. 



\subsection{Neural Machine Translation}
Using the synthetic data, we build neural MT systems for each language pair in one direction. To do so, one of the approaches is to train a transformer-based encoder-decoder model from scratch with random weights only on the parallel data. This model type excels in high-resource settings but hardly reaches up to the mark performance for low-resource languages~\cite{duh-etal-2020-benchmarking}. Another approach is to fine-tune a model based on a pre-trained model. Instead of initializing with random weights, the training is carried out on a previously-trained transformer model. The pre-trained model can be either monolingual or multilingual and can be pre-trained on any task, normally on denoising ones. This approach \cite{alabi-etal-2022-adapting} is promising to improve low-resource languages as the model does not need to learn all language components from scratch. If the pre-trained model is multilingual, the model can leverage resources from other high-resource languages.

\section{Experimental Setup}

\subsection{Dataset}

\paragraph{Parallel Data}
To create synthetic data, we use the parallel sentences in the OPUS-100~\cite{zhang-etal-2020-improving} corpus\footnote{\url{https://data.statmt.org/opus-100-corpus}} with English as the source language and other languages as target languages. We use this training set as our parallel seed data for training. For testing and validation, we use the \texttt{devtest} and \texttt{dev} sets of the FLORES-200\footnote{\url{https://github.com/facebookresearch/flores/tree/main/flores200}}  benchmark~\cite{nllbteam2022language} respectively. Table~\ref{tab:data} summarizes the statistics of our datasets. We divide the languages into four categories according to their data availability: extremely low-resource, low-resource, well-resourced, and high-resource.

\begin{table}[!t]
\centering
\begin{tabular}{r|c|c}
\toprule
\textbf{Language (code)} & \textbf{\# Seed} & \textbf{\# Entries}\\\midrule
Armenian (\textsc{hye}) & 7,059   & 161,798\\
Wolof (\textsc{wol}) & 7,918  & 4,971\\
Kurmanji (\textsc{kmr}) & 8,199  & 47,461\\
Scottish Gaelic (\textsc{gla}) & 16,316  & 51,416\\\midrule
Marathi (\textsc{mar}) & 27,007  & 65,559\\
Uyghur (\textsc{uig}) & 72,170  & 9,054\\
Kazakh (\textsc{kaz}) & 79,927 & 40,516\\\midrule
Tamil (\textsc{tam}) & 227,014 & 230,882\\
Irish (\textsc{gle}) & 289,524  & 71,436\\
Galician (\textsc{glg}) & 515,344  & 185,946\\
Hindi (\textsc{hin}) & 534,319  & 409,076\\\midrule
Urdu (\textsc{urd}) & 753,913  & 86,695\\
Greek (\textsc{ell}) & 1,000,000  & 407,311\\
Maltese (\textsc{mlt}) & 1,000,000  & 33,131\\
\bottomrule

\end{tabular}
\caption{Statistics of our datasets (seed parallel data and dictionary entries). Sorted according to the number of available training sentences.}
\label{tab:data}
\end{table}

\paragraph{Bilingual Dictionaries}
We extract dictionaries between English and each target language from the PanLex database containing 25 million words in 2,500 dictionaries of 5,700 languages.

\subsection{Tools}
We use \texttt{Stanza}\footnote{\url{https://stanfordnlp.github.io/stanza/}} for tokenization, morphological feature tagging, POS tagging, and lemmatization. Stanza uses different models for different languages. For word alignment we use \texttt{fast$\_$align}\footnote{\url{https://github.com/clab/fast_align}}~\cite{dyer-etal-2013-simple}, and we use \texttt{pyinflect}\footnote{\url{https://pypi.org/project/pyinflect/0.2.0/}} for morphological inflection. Note that \texttt{pyinflect} only supports English, but in this work, we only do inflection on the English side for our synthetic data creation framework. We use the HuggingFace~\cite{wolf2020huggingfaces} toolkit for training the language models.

\subsection{Implementation Details}
\paragraph{Language Model} To construct the language model (LM), we adopt the methodology outlined in the GPT-2 recipe provided by HuggingFace ~\cite{Radford2019LanguageMA}. We utilize the monolingual side of the parallel data specific to each language as the training dataset. Given that many of the languages involved in our experiment are not highly resourced, we make certain modifications to the GPT-2 model. We employ only six layers instead of the original 12 to mitigate resource limitations. Additionally, we decrease the vocabulary size to 5000. These adjustments help tailor the model to our experiment's specific requirements of low-resource languages.

\paragraph{Seed Data}
We do not use all the available seed data for creating synthetic sentences. Short sentences with less than seven tokens are not used as seed sentences. As Stanza uses context to generate morphological features, short sentences seem not to provide enough context for the model to produce reasonable annotations.\footnote{This was based on preliminary experiments and manual inspection of Stanza's outputs.} 

\paragraph{Lexicons}
The bilingual lexicons often provide several translations for one source word. We organize the lexicon so that only one translation is available for each source word. We do so naively, only taking the first translation of a word and discarding the rest. We also ensure that the source and target have the same POS tag. To produce the morphological features of the dictionary entries, we rely not only on Stanza\footnote{As there is no sentential context, Stanza is bound to be error-prone.} but we also perform lookups on  \emph{Unimorph}\footnote{\url{https://unimorph.github.io/}} ~\cite{batsuren-etal-2022-unimorph}, which provides morphological inflection paradigms for dozens of languages (including the ones we work on) annotated with POS tags and morphological features. For this work, we only work with augmentation, focusing on nouns, adjectives, and verbs.

\paragraph{Synthetic Data}
We create five sizes of synthetic data: 5K, 10K, 50K, 100K, and 200K for each language pair. In each set, the previous set's data is used. That means that when compiling the 10K synthetic dataset, we create new 5K data to add to the previous 5K data, and so on. This ensures that our experiments only vary based on the newly added synthetic data (and not due to additional randomness). 

For each sentence, we randomly choose at most two words for replacement. As the word replacement is random, getting the exact number of sentences for each set is not guaranteed, and there may be duplicates. From each seed sentence, $M$ number of synthetic sentences are created. Let's say we want to make a total of 5,000 seed sentences. Then, $M$ is chosen to get barely more than 5,000 sentences. After that, we sort with the perplexity of the LM and select the sentences with lower perplexity to create that set.

\paragraph{Model Details}
We fine-tune DeltaLM~\cite{DBLP:journals/corr/abs-2106-13736}, a large pre-trained multilingual encoder-decoder model that regards the decoder as the task layer of off-the-shelf pre-trained encoders. This is done separately for each language, not multilingually. The baseline system is the one that uses only the available real parallel data. Throughout the paper, we refer to the baseline as the \emph{0K (untagged)} model, as it has seen 0 synthetic data. The rest of the models are use tags \emph{<clean>} and \emph{<noisy>} at the beginning of the sentences to distinguish between real and synthetic data. The model's name (e.g.  \emph{5K}) indicates how much synthetic data has been added to the seed data during training.

\pgfplotstableread[row sep=\\,col sep=&]{
noise & ENGDNaive & ENGDOurs & ENHYNaive & ENHYOurs & ENKKNaive & ENKKOurs & ENKMRNaive & ENKMROurs & ENWONaive & ENWOOurs & ENELNaive & ENELOurs & ENGANaive & ENGAOurs & ENGLNaive & ENGLOurs & ENHINaive & ENHIOurs & ENMRNaive & ENMROurs & ENMTNaive & ENMTOurs & ENTANaive & ENTAOurs & ENURNaive & ENUROurs & ENUGNaive & ENUGOurs\\ 
40 &5.16 &4.75 &4.86 &4.28 &6.34 &6.28 &2.36 &2.45 &0.78 &1.08 &22.2 &22.17 &19.89 &20.3 &31.42 &30.97 &23.2 &23.29 &4.86 &5.47 &39.86 &40.11 &5.64 &5.44 &12.06 &12.24 &0.7 &0.63\\
60 &5.07 &5.22 &4.65 &4.52 &5.93 &6.5 &2.05 &2.27 &1.03 &1.13 &23.09 &21.93 &19.97 &19.94 &31.41 &31.27 &22.53 &23.44 &4.64 &5.35 &39.77 &39.08 &5.33 &5.76 &12.31 &12.37 &0.88 &1.06\\
80 &5.5 &5.62 &2.91 &3.3 &5.16 &5.27 &2.12 &2.32 &1.19 &1.11 &22.17 &22.24 &20.3 &20.44 &31.34 &31.13 &22.72 &22.42 &4.55 &4.76 &39.64 &39.52 &5.35 &5.28 &12.03 &11.45 &0.76 &1.17\\
100 &5.31 &5.13 &2.75 &2.97 &5.39 &5.68 &1.66 &1.89 &1.05 &1.2 &22.55 &22.64 &20.58 &20.9 &31.68 &31.46 &22.35 &22.17 &4.54 &4.72 &39.41 &38.69 &5.10 &5.48 &11.98 &11.95 &0.87 &1.28\\
120 &4.78 &4.97 &2.11 &2.83 &4.58 &4.79 &1.82 &1.78 &0.94 &1.02 &23.01 &23.07 &21.52 &22.34 & & &22.05 &21.17 &4.04 &4.34 &39.24 &39.43 &5.58 &5.34 &11.12 &11.07 & & \\
}\sinadata

\pgfplotstableread[row sep=\\,col sep=&]{
noise & ENGDNaive & ENGDOurs & ENHYNaive & ENHYOurs & ENKKNaive & ENKKOurs & ENKMRNaive & ENKMROurs & ENWONaive & ENWOOurs & ENELNaive & ENELOurs & ENGANaive & ENGAOurs & ENGLNaive & ENGLOurs & ENHINaive & ENHIOurs & ENMRNaive & ENMROurs & ENMTNaive & ENMTOurs & ENTANaive & ENTAOurs & ENURNaive & ENUROurs & ENUGNaive & ENUGOurs\\ 
140 &&&&&&&&&&&&&&&&&&&&&&&&&&&&\\
}\sinaalldata

\pgfplotsset{sina plot/.style={
            every axis plot post/.style={/pgf/number format/.cd,fixed,precision=1},
            width=5.7cm,
            height=4cm,
            ymajorgrids=false,
            yminorgrids=false,
            xtick={40,60,80,100,120},
            xticklabels={5K,10K,50K,100K,200K},
            every x tick label/.append style={font=\tiny},
            every y tick label/.append style={font=\tiny},
            ymax=100,ymin=0,
            tick pos=left,
            axis y line*=left,
            axis x line*=bottom,
            nodes near coords,
            every node near coord/.append style={font=\tiny,color=black, above},
            enlarge x limits=0.1,
            title style={yshift=-.1cm},
    }
}

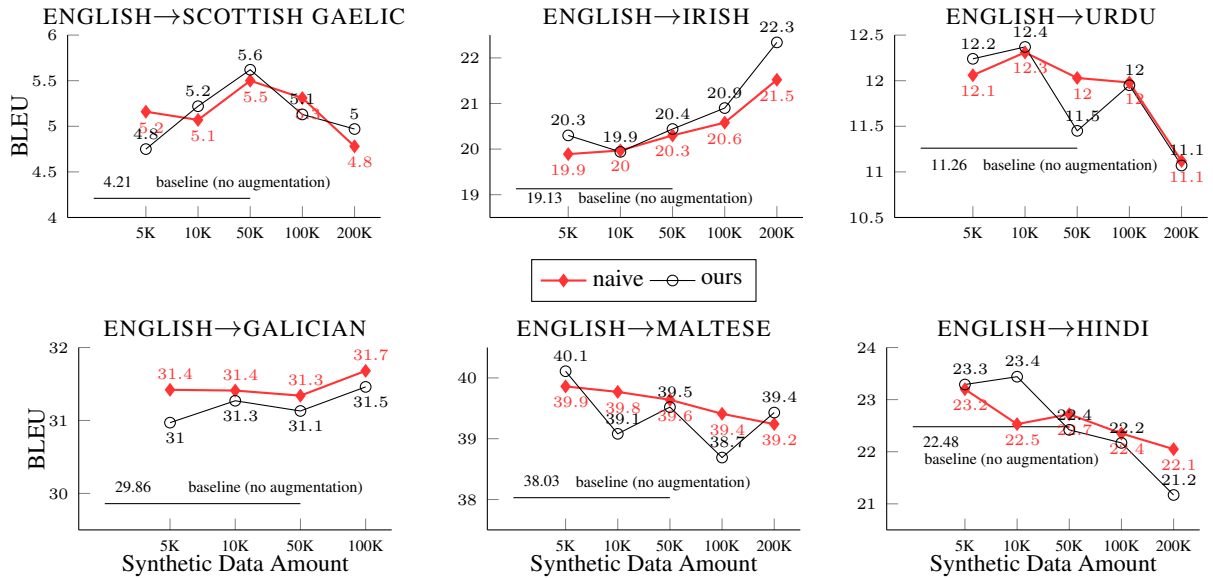
\begin{figure*}[ht]
\centering
\begin{tabular}{ccc}
    \begin{tikzpicture}
    \begin{axis}[sina plot,ymin=4, ymax=6,
    title={{\textsc{english$\rightarrow$scottish gaelic}}},
    title style={yshift=-0.1cm},
    ylabel={\small{BLEU}},
    ylabel near ticks,
    ylabel shift=-6pt,
    ]
    \addplot [mark=diamond*, every node near coord/.append style={xshift=2pt,anchor=north,color=red!80}, thick, color=red!80] table[x=noise,y=ENGDNaive]{\sinadata};
    \addplot [mark=o, every node near coord/.append style={xshift=0pt,anchor=south}] table[x=noise,y=ENGDOurs]{\sinadata};
    \newcommand\BASE{4.21}
    \addplot[mark=none, black, nodes near coords={}] coordinates {(20,\BASE) (80,\BASE)};
    \node[style={font=\tiny,color=black, anchor=west,yshift=12pt}] at  (0,\BASE) {\BASE};
    \node[style={font=\tiny,color=black, anchor=west,yshift=11.5pt}] at  (20,\BASE) {baseline (no augmentation)};
    \end{axis}
    \end{tikzpicture} 
    & 
    \begin{tikzpicture}
    \begin{axis}[sina plot, ymin=18.5, ymax=22.5,
    title={{\textsc{english$\rightarrow$irish}}},
    title style={yshift=-.1cm},
    legend style={at={(0.5,-0.2)},anchor=north,font=\small,legend columns=2},
    ]
    \addplot [mark=diamond*, every node near coord/.append style={xshift=0pt,yshift=0pt,anchor=north,color=red!80}, thick, color=red!80] table[x=noise,y=ENGANaive]{\sinadata};
    \addplot [mark=o, every node near coord/.append style={xshift=0pt,yshift=0pt,anchor=south}] table[x=noise,y=ENGAOurs]{\sinadata};
    \newcommand\BASE{19.13}
    \addplot[mark=none, black, nodes near coords={}] coordinates {(20,\BASE) (80,\BASE)};
    \node[style={font=\tiny,color=black, anchor=west,yshift=4pt}] at  (0,\BASE) {\BASE};
    \node[style={font=\tiny,color=black, anchor=west,yshift=3.5pt}] at  (20,\BASE) {baseline (no augmentation)};
    \end{axis}
    \end{tikzpicture}
    &
    \begin{tikzpicture}
    \begin{axis}[sina plot,ymin=10.5, ymax=12.5,
    title={{\textsc{english$\rightarrow$urdu}}},
    title style={yshift=-.1cm},
    legend style={at={(0.97,0.65)},anchor=south west,font=\small}
    ]
    \addplot [mark=diamond*, every node near coord/.append style={xshift=2pt,anchor=north,color=red!80}, thick, color=red!80] table[x=noise,y=ENURNaive]{\sinadata};
    \addplot [mark=o, every node near coord/.append style={xshift=2pt,anchor=south}] table[x=noise,y=ENUROurs]{\sinadata};
    \newcommand\BASE{11.26}
    \addplot[mark=none, black, nodes near coords={}] coordinates {(20,\BASE) (80,\BASE)};
    \node[style={font=\tiny,color=black, anchor=west,yshift=16pt}] at  (0,\BASE) {\BASE};
    \node[style={font=\tiny,color=black, anchor=west,yshift=15.5pt}] at  (20,\BASE) {baseline (no augmentation)};
    \end{axis}
    \end{tikzpicture}
    \\
    \begin{tikzpicture}
    \begin{axis}[sina plot, ymin=29.5, ymax=32,
    title={{\textsc{english$\rightarrow$galician}}},
    title style={yshift=-.1cm},
    ylabel={\small{BLEU}},
    ylabel near ticks,
    ylabel shift=-6pt,
    xlabel={\small{Synthetic Data Amount}},
    xlabel near ticks,
    xlabel shift=-6pt,
    legend style={at={(0.97,0.65)},anchor=south west,font=\small}
    ]
    \addplot [mark=diamond*, every node near coord/.append style={xshift=2pt,anchor=south,color=red!80}, thick, color=red!80] table[x=noise,y=ENGLNaive]{\sinadata};
    \addplot [mark=o, every node near coord/.append style={xshift=2pt,anchor=north}] table[x=noise,y=ENGLOurs]{\sinadata};
    \newcommand\BASE{29.86}
    \addplot[mark=none, black, nodes near coords={}] coordinates {(20,\BASE) (80,\BASE)};
    \node[style={font=\tiny,color=black, anchor=west,yshift=8pt}] at  (0,\BASE) {\BASE};
    \node[style={font=\tiny,color=black, anchor=west,yshift=7.5pt,xshift=25pt}] at  (20,\BASE) {baseline (no augmentation)};
    \end{axis}
    \end{tikzpicture}   
    & 
    \begin{tikzpicture}
    \begin{axis}[sina plot,ymin=37.5, ymax=40.5,
    title={{\textsc{english$\rightarrow$maltese}}},
    title style={yshift=-.1cm},
    xlabel={\small{Synthetic Data Amount}},
    xlabel near ticks,
    xlabel shift=-6pt,
    legend style={at={(0.5,1.25)},anchor=south,font=\small,legend columns=2}
    ]
    \addplot [mark=diamond*, every node near coord/.append style={xshift=2pt,anchor=north,color=red!80}, thick, color=red!80] table[x=noise,y=ENMTNaive]{\sinadata};
    \addlegendentry{naive};
    \addplot [mark=o, every node near coord/.append style={xshift=2pt,anchor=south}] table[x=noise,y=ENMTOurs]{\sinadata};
    \addlegendentry{ours};
    \newcommand\BASE{38.03}
    \addplot[mark=none, black, nodes near coords={}] coordinates {(20,\BASE) (80,\BASE)};
    \node[style={font=\tiny,color=black, anchor=west,yshift=10pt}] at  (0,\BASE) {\BASE};
    \node[style={font=\tiny,color=black, anchor=west,yshift=9pt}] at  (20,\BASE) {baseline (no augmentation)};
    \end{axis}
    \end{tikzpicture}
    &
    \begin{tikzpicture}
    \begin{axis}[sina plot, ymin=20.5, ymax=24,
    title={{\textsc{english$\rightarrow$hindi}}},
    title style={yshift=-.1cm},
    xlabel={\small{Synthetic Data Amount}},
    xlabel near ticks,
    xlabel shift=-6pt,
    legend style={at={(0.97,0.65)},anchor=south west,font=\small}
    ]
    \addplot [mark=diamond*, every node near coord/.append style={xshift=2pt,anchor=north,color=red!80}, thick, color=red!80] table[x=noise,y=ENHINaive]{\sinadata};
    \addplot [mark=o, every node near coord/.append style={xshift=2pt,anchor=south}] table[x=noise,y=ENHIOurs]{\sinadata};
    \newcommand\BASE{22.48}
    \addplot[mark=none, black, nodes near coords={}] coordinates {(20,\BASE) (80,\BASE)};
    \node[style={font=\tiny,color=black, anchor=west,yshift=29pt}] at  (0,\BASE) {\BASE};
    \node[style={font=\tiny,color=black, anchor=west,yshift=22pt,xshift=-19pt}] at  (20,\BASE) {baseline (no augmentation)};
    \end{axis}
    \end{tikzpicture}
    
\end{tabular}
\vspace*{-4mm}
\caption{BLEU scores on the test sets for six languages in the \textsc{english-x} direction. X-axis indicates the amount of synthetic parallel data we use along with seed data. The baseline uses no synthetic data. Except for Irish and Galician, all the other languages do not benefit from the increasing amounts of synthetic data. It seems like Irish has even room for more improvement. \textbf{ours} is the morphologically-informed method.}
\label{fig:En-Xresult}
\end{figure*}

\pgfplotstableread[row sep=\\,col sep=&]{
noise & HYENNaive & HYENOurs & WOENNaive & WOENOurs & ELENNaive & ELENOurs & GAENNaive & GAENOurs & GDENNaive & GDENOurs & GLENNaive & GLENOurs & HIENNaive & HIENOurs & KKENNaive & KKENOurs & KMRENNaive & KMRENOurs & MRENNaive & MRENOurs & MTENNaive & MTENOurs & TAENNaive & TAENOurs & URENNaive & URENOurs & USENNaive & UGENOurs\\ 
40 &15.78 &15.59& 2.59 &2.31 &32.33 &32.36 &30.03 &29.96 &13.05 &13.32 &37.89 &37.84 &30.89 &31.05 &20.97 &20.78 &11.63 &11.97 &23.66 &24.11 &45.2 &45.65 &20.72 &21.3 &19.95 &19.96 &11.08 &11.32\\
60 &13.84 &14.63 &2.24 &2.22 &32.32 &31.88 &29.54 &29.7 &11.81 &12.76 &38.01 &38.19 &30.34 &30.74 &20.74 &21.3 &11.08 &12.01 &23.36 &23.54 &45.43 &45.07 &20.61 &20.69 &19.81 &19.39 &11.27 &11.31\\
80 &9.71 &10.11 &2.09 &1.71 &32.34 &32.34 &28.65 &30.03 &12.99 &12.32 &37.89 &38.02 &30.81 &30.58 &18.86 &20.52 &11.04 &10.87 &12.1 &12.05 &45.23 &44.87 &19.9 &19.75 &19.75 &19.15 &10.67 &10.71\\
100 &8.39 &9.24 &1.53 &71.55 &32.35 &32.23 &28.73 &28.99 &12.36 &13.08 &38.02 &37.61 &30.46 &29.6 &17.99 &19.28 &9.45 &10.35 &21.22 &21.78 &45.23 &44.94 &18.98 &19.25 &19.00 &19.44 &9.56 &9.76\\
120 &7.79 &8.72 &1.21 &1.1 &31.98 &32.1 &28.57 &29.36 &11.86 &12.35 & & &29.71 &29.62 &16.59 &18.45 &9.19 &9.43 &20.00 &19.92 &44.90 &44.7 &18.11 &18.63 &18.12 &18.54 &8.77 &8.18\\
}\sinadataxen

\pgfplotstableread[row sep=\\,col sep=&]{
noise & HYENNaive & HYENOurs & WOENNaive & WOENOurs & ELENNaive & ELENOurs & GAENNaive & GAENOurs & GDENNaive & GDENOurs & GLENNaive & GLENOurs & HIENNaive & HIENOurs & KKENNaive & KKENOurs & KMRENNaive & KMRENOurs & MRENNaive & MRENOurs & MTENNaive & MTENOurs & TAENNaive & TAENOurs & URENNaive & URENOurs & USENNaive & UGENOurs\\  
140 &&&&&&&&&&&&&&&&&&&&&&&&&&&&\\
}\sinaalldataxen

\pgfplotsset{sina plot/.style={
            every axis plot post/.style={/pgf/number format/.cd,fixed,precision=1},
            width=5.7cm,
            height=4cm,
            ymajorgrids=false,
            yminorgrids=false,
            xtick={40,60,80,100,120},
            xticklabels={5K,10K,50K,100K,200K},
            every x tick label/.append style={font=\tiny},
            every y tick label/.append style={font=\tiny},
            ymax=100,ymin=0,
            tick pos=left,
            axis y line*=left,
            axis x line*=bottom,
            nodes near coords,
            every node near coord/.append style={font=\tiny,color=black, above},
            enlarge x limits=0.1,
            title style={yshift=-.1cm},
    }
}

\begin{figure*}[ht]
\centering
\begin{tabular}{ccc}
    \begin{tikzpicture}
    \begin{axis}[sina plot,ymin=8.5, ymax=14,
    title={{\textsc{scottish gaelic$\rightarrow$english}}},
    title style={yshift=-0.1cm},
    ylabel={\small{BLEU}},
    ylabel near ticks,
    ylabel shift=-6pt,
    ]
    \addplot [mark=diamond*, every node near coord/.append style={xshift=2pt,anchor=north,color=red!80}, thick, color=red!80] table[x=noise,y=GDENNaive]{\sinadataxen};
    \addplot [mark=o, every node near coord/.append style={xshift=0pt,anchor=south}] table[x=noise,y=GDENOurs]{\sinadataxen};
    \newcommand\BASE{9.08}
    \addplot[mark=none, black, nodes near coords={}] coordinates {(20,\BASE) (80,\BASE)};
    \node[style={font=\tiny,color=black, anchor=west,yshift=12pt}] at  (0,\BASE) {\BASE};
    \node[style={font=\tiny,color=black, anchor=west,yshift=11.5pt}] at  (20,\BASE) {baseline (no augmentation)};
    \end{axis}
    \end{tikzpicture} 
    & 
    \begin{tikzpicture}
    \begin{axis}[sina plot, ymin=28.5, ymax=31.5,
    title={{\textsc{hindi$\rightarrow$english}}},
    title style={yshift=-.1cm},
    legend style={at={(0.5,-0.2)},anchor=north,font=\small,legend columns=2},
    ]
    \addplot [mark=diamond*, every node near coord/.append style={xshift=0pt,yshift=0pt,anchor=north,color=red!80}, thick, color=red!80] table[x=noise,y=HIENNaive]{\sinadataxen};
    \addplot [mark=o, every node near coord/.append style={xshift=0pt,yshift=0pt,anchor=south}] table[x=noise,y=HIENOurs]{\sinadataxen};
    \newcommand\BASE{29.09}
    \addplot[mark=none, black, nodes near coords={}] coordinates {(20,\BASE) (80,\BASE)};
    \node[style={font=\tiny,color=black, anchor=west,yshift=3pt}] at  (0,\BASE) {\BASE};
    \node[style={font=\tiny,color=black, anchor=west,yshift=2.5pt}] at  (20,\BASE) {baseline (no augmentation)};
    \end{axis}
    \end{tikzpicture}
    &
    \begin{tikzpicture}
    \begin{axis}[sina plot,ymin=16, ymax=21.5,
    title={{\textsc{kazakh$\rightarrow$english}}},
    title style={yshift=-.1cm},
    legend style={at={(0.97,0.65)},anchor=south west,font=\small}
    ]
    \addplot [mark=diamond*, every node near coord/.append style={xshift=2pt,anchor=north,color=red!80}, thick, color=red!80] table[x=noise,y=KKENNaive]{\sinadataxen};
    \addplot [mark=o, every node near coord/.append style={xshift=2pt,anchor=south}] table[x=noise,y=KKENOurs]{\sinadataxen};
    \newcommand\BASE{19.58}
    \addplot[mark=none, black, nodes near coords={}] coordinates {(20,\BASE) (80,\BASE)};
    \node[style={font=\tiny,color=black, anchor=west,yshift=38pt}] at  (0,\BASE) {\BASE};
    \node[style={font=\tiny,color=black, anchor=west,yshift=32pt},xshift=-19pt] at  (20,\BASE) {baseline (no augmentation)};
    \end{axis}
    \end{tikzpicture}
    \\
    \begin{tikzpicture}
    \begin{axis}[sina plot, ymin=9, ymax=12.5,
    title={{\textsc{kurmanji$\rightarrow$english}}},
    title style={yshift=-.1cm},
    ylabel={\small{BLEU}},
    ylabel near ticks,
    ylabel shift=-6pt,
    xlabel={\small{Synthetic Data Amount}},
    xlabel near ticks,
    xlabel shift=-6pt,
    legend style={at={(0.97,0.65)},anchor=south west,font=\small}
    ]
    \addplot [mark=diamond*, every node near coord/.append style={xshift=2pt,anchor=south,color=red!80}, thick, color=red!80] table[x=noise,y=KMRENNaive]{\sinadataxen};
    \addplot [mark=o, every node near coord/.append style={xshift=2pt,anchor=north}] table[x=noise,y=KMRENOurs]{\sinadataxen};
    \newcommand\BASE{9.73}
    \addplot[mark=none, black, nodes near coords={}] coordinates {(20,\BASE) (80,\BASE)};
    \node[style={font=\tiny,color=black, anchor=west,yshift=8pt}] at  (0,\BASE) {\BASE};
    \node[style={font=\tiny,color=black, anchor=west,yshift=7.5pt,xshift=-5pt}] at  (20,\BASE) {baseline (no augmentation)};
    \end{axis}
    \end{tikzpicture}   
    & 
    \begin{tikzpicture}
    \begin{axis}[sina plot,ymin=17.5, ymax=21.5,
    title={{\textsc{tamil$\rightarrow$english}}},
    title style={yshift=-.1cm},
    xlabel={\small{Synthetic Data Amount}},
    xlabel near ticks,
    xlabel shift=-6pt,
    legend style={at={(0.5,1.25)},anchor=south,font=\small,legend columns=2}
    ]
    \addplot [mark=diamond*, every node near coord/.append style={xshift=2pt,anchor=north,color=red!80}, thick, color=red!80] table[x=noise,y=TAENNaive]{\sinadataxen};
    \addlegendentry{naive};
    \addplot [mark=o, every node near coord/.append style={xshift=2pt,anchor=south}] table[x=noise,y=TAENOurs]{\sinadataxen};
    \addlegendentry{ours};
    \newcommand\BASE{19.83}
    \addplot[mark=none, black, nodes near coords={}] coordinates {(20,\BASE) (80,\BASE)};
    \node[style={font=\tiny,color=black, anchor=west,yshift=32pt}] at  (0,\BASE) {\BASE};
    \node[style={font=\tiny,color=black, anchor=west,yshift=25pt,xshift=-19pt}] at  (20,\BASE) {baseline (no augmentation)};
    \end{axis}
    \end{tikzpicture}
    &
    \begin{tikzpicture}
    \begin{axis}[sina plot, ymin=17.5, ymax=20.5,
    title={{\textsc{urdu$\rightarrow$english}}},
    title style={yshift=-.1cm},
    xlabel={\small{Synthetic Data Amount}},
    xlabel near ticks,
    xlabel shift=-6pt,
    legend style={at={(0.97,0.65)},anchor=south west,font=\small}
    ]
    \addplot [mark=diamond*, every node near coord/.append style={xshift=2pt,anchor=north,color=red!80}, thick, color=red!80] table[x=noise,y=URENNaive]{\sinadataxen};
    \addplot [mark=o, every node near coord/.append style={xshift=2pt,anchor=south}] table[x=noise,y=URENOurs]{\sinadataxen};
    \newcommand\BASE{18.63}
    \addplot[mark=none, black, nodes near coords={}] coordinates {(20,\BASE) (80,\BASE)};
    \node[style={font=\tiny,color=black, anchor=west,yshift=16pt}] at  (0,\BASE) {\BASE};
    \node[style={font=\tiny,color=black, anchor=west,yshift=15.5pt,xshift=2pt}] at  (20,\BASE) {baseline (no augmentation)};
    \end{axis}
    \end{tikzpicture}
    
\end{tabular}
\vspace*{-4mm}
\caption{BLEU scores on the test sets for six languages in the \textsc{x-English} direction. \textbf{ours} is the morphologically-informed method. Morphologically-informed approach outperforms the naive approach in all the six language pairs. X-axis indicates the amount of synthetic parallel data we use along with seed data. The baseline uses no synthetic data. }
\label{fig:X-Enresult}
\end{figure*}
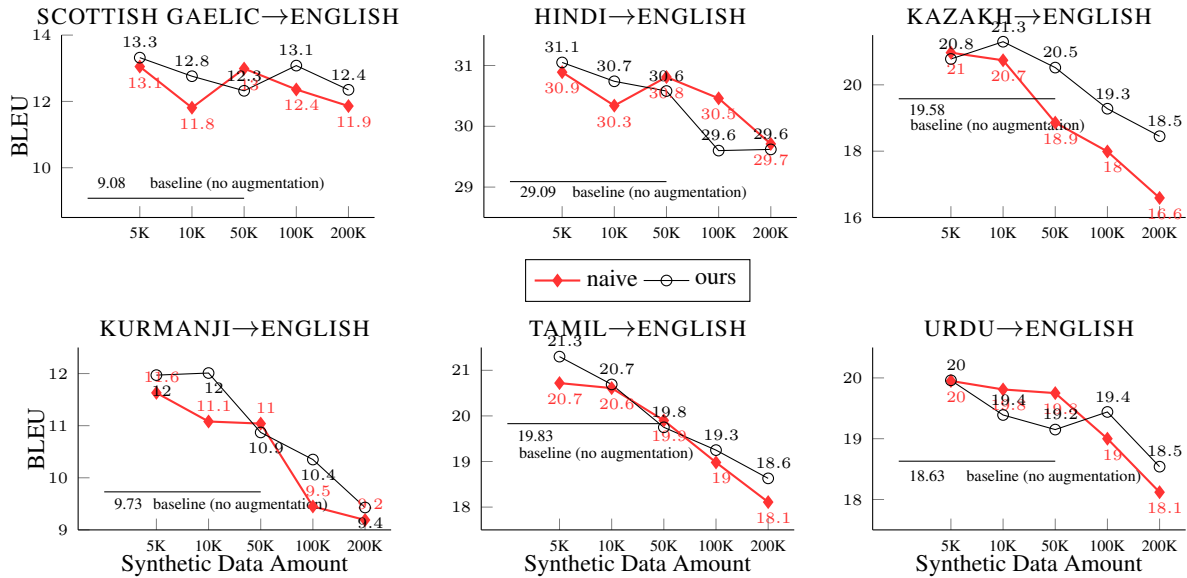

\section{Results}
\paragraph{From English}
In 11 out of the 14 Eng-X language pairs, our approaches yield improvements randgng from 0.4 points to more than 3 BLEU~\cite{papineni-etal-2002-bleu} points. 

Due to space constraint, we show all results for the six language pairs with the largest improvement over the baseline in Figure~\ref{fig:En-Xresult}. 
Comparing the data augmentation methods, our morphologically aware approach yields a better score than the naive one in all cases except for Galician. We find that the augmentation is consistently beneficial for Irish and Galician, regardless of how much data we add. But for other pairs adding more synthetic data does not lead to sustained improvements.

Table~\ref{tab:main_table1} in the Appendix shows our experimental results on all 14 language pairs from English. We use two pre-trained models: \emph{DeltaLM-Base} and \emph{DeltaLM-Large}. We tried to use DeltaLM-Large for all language pairs, but low-resource language pairs quickly overfit on the large model and do not generalize well. Apart from Armenian (\textsc{hye}), we get a higher BLEU score with our settings for all other language pairs. The improvement margin is negligent in languages where the baseline system is already very bad. Languages like Wolof (\textsc{wol}) and Uyghur (\textsc{uig}) have baseline BLEU scores of less than 2, showing us that our parallel seed data is not of good quality. 
For languages, Kazakh (\textsc{kaz}), Marathi (\textsc{mar}), and Tamil (\textsc{tam}), all rather low-resource languages, the improvement is less than 0.5 BLEU points, but it ranges from 0.79 to 3.21 BLEU score for all other languages. We also observe a similar trend of improvement in the case of adding more noisy data. The score improves to the highest point, but as more synthetic data is added the system gets worse.

\paragraph{To English}
As before we show the best-performing six language pairs in the X-English direction in Figure~\ref{fig:X-Enresult}. Unlike English-X, the patterns here are the same for all language pairs. In all 14 languages except for Armenian our approach improves upon the baseline, and the morphologically-informed method is better than the naive approach. In every case adding more synthetic data after a while does not lead to more improvements.

Table~\ref{tab:main_table2} in the Appendix lists our results on all 14 language pairs in the to-English direction. The BLEU scores are generally higher in this setting, as the pre-trained model has seen a lot of English data on the target side. Apart from Armenian (\textsc{hye}), we get a higher BLEU score with our settings for all other language pairs, the same as before. The improvement margin is negligent for Greek (\textsc{ell}) and Maltese (\textsc{mlt}), showing that the language has no room for improvement through this type of augmentation, as the models are already fairly good. For Wolof (\textsc{wol}) and Uyghur (\textsc{uig}), our improvement is less than 1 BLEU score. This is also the same as before, showing that the parallel data for these languages is not of high quality. The improvement ranges from 1.12 to 4.24 BLEU scores for all other languages. We also see a similar trend as in the From-English direction; after some point, the more synthetic data we add, the system worsens; most improvements are obtained with 5K and 10K synthetic examples.

\section{Analysis}
\paragraph{Is the performance tied to any single component?}
We perform an ablation study to find out which component of the model is responsible for the performance boost. We work on this experiment with five thousand synthetic data and the \textsc{scottish gaelic-english} direction. We compare three different components of our pipeline:

\begin{itemize}[noitemsep,leftmargin=*]
    \item \textbf{Does the number of the generated synthetic data matter?} For this setup, instead of creating three synthetic sentences from each seed sentence, we create 30 synthetic sentences from each seed sentence. We refer to this setup as a ``5K Number''.
    \item \textbf{Does the length of the seed sentences matter?} We create synthetic sentences from seed sentences with less than seven tokens for this setup. We refer to this setup as a ``5K Length''.
    \item \textbf{Does the choice of alignment model play any role?} Instead of aligning the seed sentences with \texttt{fast$\_$align}, we use \texttt{awesome-align} using ``bert-base-multilingual-cased'' for this setup. We refer to this setup as a ``5K Align''.
\end{itemize}

\begin{table}[!t]
\centering
\begin{tabular}{r|c|c}
\toprule
\multicolumn{3}{c}{\textbf{Ablations on Scottish Gaelic-English}}  \\ 
\midrule
             & \textbf{Ours}  & \textbf{Naive}        \\ 
\midrule
5K           & \textbf{13.32} & 13.05        \\
5K Number  & 12.76 & \textbf{13.13}        \\
5K Length & 12.58 & 12.28        \\
5K Align   & 12.14 & 12.62        \\
\bottomrule
\end{tabular}

\caption{Ablation result of the importance of different components of our method. If we don't use one of the components, the BLEU score drops significantly.}
\label{tab:abl}
\end{table}


Table~\ref{tab:abl} shows the results of these experiments. The main takeaway is that every component of our method is necessary to boost scores: scores decrease when we replace one component. The most important for low-resource languages is to use a compatible alignment model. As large multilingual pre-trained models do not represent them very well, and \texttt{awesome\_align} relies on such a model, we are better off using \texttt{fast\_align}. The number of generated synthetic data also matters as we anticipated. The reason is that when we create a huge sentence pool and sample a small number of sentences from there, the number of unique seed sentences that contribute to the synthetic data is reduced. We also confirm that the length of the sentence matters for Stanza: the shorter the sentence is, the less context it has, thus reducing the quality of the morphological analysis and consequently of our synthetic sentences.

\paragraph{Does the number of seed sentences or the amount of new vocabulary matter?}
To do this experiment, we work again on the GLA-ENG direction to create five thousand synthetic data. We build four different models:

\begin{itemize}[noitemsep]
    \item \textbf{5K}: This is the original five thousand-size synthetic dataset we created. We create three sentences from each seed sentence and randomly choose words from all candidate words for replacement.
    \item \textbf{5K (one)}: In this setup, we try to create 5000 sentences from only one seed sentence and randomly choose words from all candidate words for replacement. However, our process could not generate 5,000 unique synthetic sentences from one seed sentence; instead, it took five seed sentences to generate 5,000 synthetic sentences.
    \item \textbf{5K (half)}: In this setup, we create three sentences from each seed sentence and randomly choose words from the first half of the candidate words for replacement.
    \item \textbf{5K (remove)}: In this setup, we create ten sentences from each seed sentence and randomly choose words from all candidate words for replacement, but when we choose a word, we remove that word as a candidate so that it is not chosen again. Ideally, we would have sentences of the same amount of the lexicon vocabulary.
\end{itemize}

\begin{table*}[!t]\centering
\small
\begin{tabular}{r|c|ccc|c|ccc}
\toprule
& \multicolumn{4}{c|}{\textbf{Scottish Gaelic-English Ours}} & \multicolumn{4}{|c}{\textbf{Scottish Gaelic-English Naive}} \\
&  & \multirow{1}{*}{\# ENG} & \multirow{1}{*}{\# GLA} & \multirow{1}{*}{\# seed} &  & \multirow{1}{*}{\# ENG} & \multirow{1}{*}{\# GLA} & \multirow{1}{*}{\# seed}\\
& BLEU & types & types & sentences & BLEU & types & types & sentences \\\midrule
0K &9.08 &11077 &13826 &0 &9.08 &11077 &13836 &0 \\
5K (one) &12.79 &11269 &14020 &5 &12.58 &12588 &15883 &5 \\
5K &\textbf{13.32} &12763 &15847 &1511 &13.05 &12057 &15082 &1512 \\
5K (half) &13.13 &12367 &15216 &1527 &12.57 &11811 &14740 &1508 \\
5K (remove) &12.91 &12528 &15702 &1909 &\textbf{13.29} &12511 &15694 &1909 \\\bottomrule
\end{tabular}
\caption{Ablation about the number of new vocabulary introduced and the number of seed sentences used to create five thousand synthetic data. Using as little as five seed sentences boosts the 3.71 BLEU score.}
\label{tab:abl_1}
\end{table*}

Table~\ref{tab:abl_1} shows these experiments' results. An exciting result is the one for 5K (one), where we use only five seed sentences to create five thousand synthetic sentences. In doing so, we introduce 200 new words, but we get a substantial jump of 3.71 BLEU score, which shows the promise of our method. Even if we have few high-quality parallel sentences and a good-quality lexicon, our method is bound to boost MT system quality.

\paragraph{How much does filtering with LM help?}
To do this experiment, we perform a control experiment to contrast with our filtering-with-LM-perplexity approach. In the control setting, we choose sentences randomly from the pool of synthetic sentences. We randomly select a subset of a hundred thousand seed sentences from the OPUS-100 ENG, ELL dataset and do ablation on both ENG-ELL and ELL-ENG directions.

\begin{table*}[!t]
\centering
\begin{tabular}{c|cccccc}
\toprule
\textbf{Pairs} &\textbf{0K } &\textbf{5K } &\textbf{10K } &\textbf{50K } &\textbf{100K} &\textbf{200K} \\\midrule
English-Modern Greek (random) &\multirow{2}{*}{14.24} &13.98 &5.01 &\textbf{15.7} &14.92 &4.23 \\
English-Modern Greek (filtered) & &14.62 &15.14 &15.2 &14.99 &\textbf{15.54} \\\midrule
Modern Greek-English (random) &\multirow{2}{*}{11.6} &12.2 &10.24 &\textbf{18.34} &9.21 &12.42 \\
Modern Greek-English (filtered) & &16.91 &17.25 &17.05 &\textbf{19.01} &17.64 \\\bottomrule
\end{tabular}
\caption{Filtering the synthetic data leads to consistent improvements, but random data sampling leads to unstable results. Instead, the BLEU score drops randomly for a random approach.}
\label{tab:abl_2}
\end{table*}

Table~\ref{tab:abl_2} shows these experiments' results. In the random setting, the results are rather unstable, with very low BLEU scores for some settings. This could be because we might be randomly choosing bad sentences from the pool. The results with informed sentence selection (through perplexity), instead, are stable and consistently improving.

\section{Related Work}
Dictionaries have been and are indispensable resources in various applications in NLP ~\cite{wilson2020urban,wang2019incorporating,DBLP:conf/conll/XiaoG14}. More specifically, many previous studies use dictionaries in MT to improve translation quality for low-resource languages with or without monolingual or parallel corpora. A closely related task is bilingual lexicon induction that departs from an unsupervised MT task where no parallel resources, including the ground-truth bilingual lexicon, are incorporated ~\cite{DBLP:journals/corr/abs-1710-11041, lample-etal-2018-phrase}. The bilingual lexicon is often utilized as a seed in bilingual lexicon induction that aims to induce more word pairs within the language pair ~\cite{DBLP:journals/corr/MikolovLS13}. Another utilization of the bilingual lexicon is for translating low-frequency words in supervised neural MT ~\cite{arthur-etal-2016-incorporating, DBLP:journals/corr/ZhangZ16c}. 

On the usage of dictionaries in MT, ~\citet{DBLP:journals/corr/abs-2004-02577} employ dictionaries for cross-lingual MT, ~\citet{fadaee-etal-2017-data} propose a data augmentation approach to target low-frequency words by generating sentence pairs containing rare words, ~\citet{DBLP:conf/acl/DuanJJTZCLZ20} use dictionaries to drive the semantic spaces of the source and target languages becoming closer in MT training without parallel sentences and ~\citet{wang2022expanding} explore the utilization of dictionaries for synthesizing textual or labeled data, focusing on tasks such as named entity recognition and part-of-speech tagging.

Unlike many of the previous approaches that are fixated on only monolingual data, our approach considers using a bilingual lexicon and maintaining morphology in augmentation. Our approach is similar in spirit to ~\citet{fadaee-etal-2017-data} technique with additional consideration of morphological complexity in the synthetic data augmentation process. Also, inspired by ~\citet{wang2022expanding}'s approach, our research shares a common thread by using different strategies for synthesizing data using lexicons and integrating such data with monolingual or parallel text when accessible. Both studies aim to leverage lexicons to enhance various NLP tasks, albeit in different contexts.

\section{Conclusion}
Our approaches have proven beneficial for most of the 14 languages under investigation, except for Armenian. Even if the improvements in BLEU scores may be small for some languages, there is a noticeable boost in most. Interestingly, we observed improvements even in language pairs (e.g., \textsc{wol-eng, eng-kmr, eng-gla}) with unsatisfactory initial baseline scores. This observation suggests our approach can enhance performance even in more challenging scenarios. The results also highlight the importance of obtaining high-quality seed sentences. We found that as few as five good-quality seed data points can contribute to creating five thousand synthetic data samples of good quality that would boost performance. This data augmentation process could play a vital role in improving the overall performance of machine translation systems and be combined with other augmentation techniques (e.g., back-translation) as we deem it orthogonal to them.\footnote{We note that back-translation is rarely effective in most extremely low-resource languages due to the abysmal quality of the initial systems.}

\paragraph{Future Work}
In our current work, we focused on conducting morphological inflection exclusively on the English side of the translation task. The main reason for this choice was the availability of a reliable morphological inflector specifically designed for English. However, we encountered challenges when applying the same approach to other languages. We lacked suitable morphological inflection tools for those languages, or the accuracy of the available tools did not meet our requirements. Incorporating these tools would have posed a significant bottleneck to the effectiveness of our approach. For future research, we aim to explore how our approach can be extended by performing morphological inflection in other languages. This involves developing or obtaining accurate and reliable morphological inflection tools.

\section{Limitations}
One of the limitations of our current approach is the use of the Stanza model. Since bilingual lexicons have no context, relying solely on morphological features obtained from the lexicon results in more general features. This can be particularly challenging in morphologically rich languages, where a single word can have multiple meanings depending on the sentence context. Another limitation is the language support provided by the Stanza model, which is currently limited to 60 languages. This constraint restricts the applicability of our approach to only those languages supported by Stanza. To expand our work to languages not supported by Stanza, it is necessary to create custom Stanza models specifically tailored for those languages. This process requires additional time and effort to develop and validate the models for each language of interest.



\bibliography{anthology,custom}


\onecolumn
\appendix

\counterwithin{figure}{section}
\counterwithin{table}{section}

\section{Appendix}
\label{sec:appendix}

\begin{table}[htb]
\centering
\small
\begin{tabular}{c@{ }c@{ }c@{ }c@{ }c@{ }c@{ }c@{ }c@{ }}
\textbf{Pairs} &\textbf{0K (untagged)} &\textbf{5K (tagged)} &\textbf{10K (tagged)} &\textbf{50K (tagged)} &\textbf{100K (tagged)} &\textbf{200K (tagged)} &$\Delta$\\\midrule
\multicolumn{7}{c}{DeltaLM-Base} \\\midrule
ENG-GLA Naive &\multirow{2}{*}{4.21} &5.16 &5.07 &5.5 &5.31 &4.78 &\multirow{2}{*}{1.41}\\
ENG-GLA Ours & &4.75 &5.22 &\textbf{5.62} &5.13 &4.97 &\\\midrule
ENG-HYE Naive &\textbf{\multirow{2}{*}{5.61}} &4.86 &4.65 &2.91 &2.75 &2.11 &\multirow{2}{*}{0.0}\\
ENG-HYE Ours & &4.28 &4.52 &3.3 &2.97 &2.83 \\\midrule
ENG-KAZ Naive &\multirow{2}{*}{6.13} &6.34 &5.93 &5.16 &5.39 &4.58 &\multirow{2}{*}{0.37}\\
ENG-KAZ Ours &&6.28 &\textbf{6.5} &5.27 &5.68 &4.79 \\\midrule
ENG-KMR Naive &\multirow{2}{*}{1.66} &2.36 &2.05 &2.12 &1.66 &1.82 &\multirow{2}{*}{0.79}\\
ENG-KMR Ours & &\textbf{2.45} &2.27 &2.32 &1.89 &1.78 \\\midrule
ENG-WOL Naive &\multirow{2}{*}{1.08} &0.78 &1.03 &1.19 &1.05 &0.94 &\multirow{2}{*}{0.12}\\
ENG-WOL Ours & &1.08 &1.13 &1.11 &\textbf{1.2} &1.02 \\\midrule
\multicolumn{7}{c}{DeltaLM-Large} \\\midrule
ENG-ELL Naive &\multirow{2}{*}{23.06} &22.2 &\textbf{23.09} &22.17 &22.55 &23.01 &\multirow{2}{*}{0.03}\\
ENG-ELL Ours & &22.17 &21.93 &22.24 &22.64 &23.07 \\\midrule
ENG-GLE Naive &\multirow{2}{*}{19.13} &19.89 &19.97 &20.3 &20.58 &21.52 &\multirow{2}{*}{3.21}\\
ENG-GLE Ours & &20.3 &19.94 &20.44 &20.9 &\textbf{22.34} \\\midrule
ENG-GLG Naive &\multirow{2}{*}{29.86} &31.42 &31.41 &31.34 &\textbf{31.68} & &\multirow{2}{*}{1.82}\\
ENG-GLG Ours & &30.97 &31.27 &31.13 &31.46 & \\\midrule
ENG-HIN Naive &\multirow{2}{*}{22.48} &23.2 &22.53 &22.72 &22.35 &22.05 &\multirow{2}{*}{0.96}\\
ENG-HIN Ours & &23.29 &\textbf{23.44} &22.42 &22.17 &21.17 \\\midrule
ENG-MAR Naive &\multirow{2}{*}{5.03} &4.86 &4.64 &4.55 &4.54 &4.04 &\multirow{2}{*}{0.44}\\
ENG-MAR Ours & &\textbf{5.47} &5.35 &4.76 &4.72 &4.34 \\\midrule
ENG-MLT Naive &\multirow{2}{*}{38.03} &39.86 &39.77 &39.64 &39.41 &39.24 &\multirow{2}{*}{2.08}\\
ENG-MLT Ours & &\textbf{40.11} &39.08 &39.52 &38.69 &39.43 \\\midrule
ENG-TAM Naive &\multirow{2}{*}{5.3} &5.64 &5.33 &5.35 &5.10 &5.58 &\multirow{2}{*}{0.46}\\
ENG-TAM Ours & &5.44 &\textbf{5.76} &5.28 &5.48 &5.34 \\\midrule
ENG-URD Naive &\multirow{2}{*}{11.26} &12.06 &12.31 &12.03 &11.98 &11.12 &\multirow{2}{*}{1.11}\\
ENG-URD Ours & &12.24 &\textbf{12.37} &11.45 &11.95 &11.07 \\\midrule
ENG-UIG Naive &\multirow{2}{*}{1.24} &0.7 &0.88 &0.76 &0.87 & &\multirow{2}{*}{0.04}\\
ENG-UIG Ours & &0.63 &1.06 &1.17 &\textbf{1.28} & \\\bottomrule
\end{tabular}
\caption{BLEU score of 9 languages from ENG-X direction. Columns indicate the amount of synthetic parallel data we use. The \textbf{0K (untagged)} column is our baseline. The rows indicating \textbf{Naive} is the approach where we replace words of the same POS tag. The rows indicating \textbf{Ours} is the approach where we replace words of the same morphological feature. $\Delta$ is the difference between the baseline and the best model's score. $\Delta$ is 0.0 if the baseline is the best model.
}
\label{tab:main_table1}
\end{table}

\begin{table}[!t]\centering
\small
\begin{tabular}{c@{ }c@{ }c@{ }c@{ }c@{ }c@{ }c@{ }c@{ }}
\textbf{Pairs} &\textbf{0K (untagged)} &\textbf{5K (tagged)} &\textbf{10K (tagged)} &\textbf{50K (tagged)} &\textbf{100K (tagged)} &\textbf{200K (tagged)} &$\Delta$\\\midrule
\multicolumn{7}{c}{DeltaLM-Base} \\\midrule
HYE-ENG Naive &\textbf{\multirow{2}{*}{16.04}} &15.78 &13.84 &9.71 &8.39 &7.79 &\multirow{2}{*}{0.0}\\
HYE-ENG Ours & &15.59 &14.63 &10.11 &9.24 &8.72 &\\\midrule
WOL-ENG Naive &\multirow{2}{*}{1.83} &\textbf{2.59} &2.24 &2.09 &1.53 &1.21 &\multirow{2}{*}{0.76}\\
WOL-ENG Ours & &2.31 &2.22 &1.71 &1.55 &1.1 \\\midrule
\multicolumn{7}{c}{DeltaLM-Large} \\\midrule
GLA-ENG Naive &\multirow{2}{*}{9.08} &13.05 &11.81 &12.99 &12.36 &11.86 &\multirow{2}{*}{4.24}\\
GLA-ENG Ours & &\textbf{13.32} &12.76 &12.32 &13.08 &12.35 \\\midrule
KAZ-ENG Naive &\multirow{2}{*}{19.58} &20.97 &20.74 &18.86 &17.99 &16.59 &\multirow{2}{*}{1.72}\\
KAZ-ENG Ours & &20.78 &\textbf{21.3} &20.52 &19.28 &18.45 \\\midrule
KMR-ENG Naive &\multirow{2}{*}{9.73} &11.63 &11.08 &11.04 &9.45 &9.19 &\multirow{2}{*}{2.28}\\
KMR-ENG Ours & &11.97 &\textbf{12.01} &10.87 &10.35 &9.43 \\\midrule
ELL-ENG Naive &\multirow{2}{*}{31.94} &32.33 &32.32 &32.34 &32.35 &31.98 &\multirow{2}{*}{0.42}\\
ELL-ENG Ours & &\textbf{32.36} &31.88 &32.34 &32.23 &32.1 \\\midrule
GLE-ENG Naive &\multirow{2}{*}{28.71} &30.03 &29.54 &28.65 &28.73 &28.57 &\multirow{2}{*}{1.32}\\
GLE-ENG Ours & &29.96 &29.7 &\textbf{30.03} &28.99 &29.36 \\\midrule
GLG-ENG Naive &\multirow{2}{*}{37.07} &37.89 &38.01 &37.89 &38.02 & &\multirow{2}{*}{1.12}\\
GLG-ENG Ours & &37.84 &\textbf{38.19} &38.02 &37.61 & \\\midrule
HIN-ENG Naive &\multirow{2}{*}{29.09} &30.89 &30.34 &30.81 &30.46 &29.71 &\multirow{2}{*}{1.96}\\
HIN-ENG Ours & &\textbf{31.05} &30.74 &30.58 &29.6 &29.62 \\\midrule
MAR-ENG Naive &\multirow{2}{*}{22.96} &23.66 &23.36 &22.1 &21.22 &20 &\multirow{2}{*}{1.15}\\
MAR-ENG Ours & &\textbf{24.11} &23.54 &22.05 &21.78 &19.92 \\\midrule
MLT-ENG Naive &\multirow{2}{*}{45.21} &45.2 &45.43 &45.23 &45.23 &44.9 &\multirow{2}{*}{0.44}\\
MLT-ENG Ours & &\textbf{45.65} &45.07 &44.87 &44.94 &44.7 \\\midrule
TAM-ENG Naive &\multirow{2}{*}{19.83} &20.72 &20.61 &19.9 &18.98 &18.11 &\multirow{2}{*}{1.47}\\
TAM-ENG Ours & &\textbf{21.3} &20.69 &19.75 &19.25 &18.63 \\\midrule
URD-ENG Naive &\multirow{2}{*}{18.63} &19.95 &19.81 &19.75 &19 &18.12 &\multirow{2}{*}{1.33}\\
URD-ENG Ours & &\textbf{19.96} &19.39 &19.15 &19.44 &18.54 \\\midrule
UIG-ENG Naive &\multirow{2}{*}{10.49} &11.08 &11.27 &10.67 &9.56 &8.77 &\multirow{2}{*}{0.83}\\
UIG-ENG Ours & &\textbf{11.32} &11.31 &10.71 &9.76 &8.18 \\\bottomrule
\end{tabular}
\caption{BLEU score of 9 languages from X-ENG direction. Columns indicate the amount of synthetic parallel data we use. The \textbf{0K (untagged)} column is our baseline. The rows indicating \textbf{Naive} is the approach where we replace words of the same POS tag. The rows indicating \textbf{Ours} is the approach where we replace words of the same morphological features. $\Delta$ is the difference between the baseline and the best model's score. $\Delta$ is 0.0 if the baseline is the best model.
}
\label{tab:main_table2}
\end{table}

\end{document}